%% file: main-6291-Aggazzotti.tex
\documentclass[11pt,a4paper]{article}
\usepackage{times,latexsym}
\usepackage{url}
\usepackage[T1]{fontenc}
\usepackage{booktabs}
\usepackage{stfloats}
\usepackage{multirow}
\usepackage{subcaption}
\usepackage{graphicx}

\usepackage{xspace,mfirstuc,tabulary}

\usepackage[acceptedWithA]{tacl2021v1}
\newif\iftaclinstructions
\taclinstructionsfalse 
\iftaclinstructions
\renewcommand{\confidential}{}
\renewcommand{\anonsubtext}{(No author info supplied here, for consistency with
TACL-submission anonymization requirements)}
\newcommand{\instr}
\fi

\iftaclpubformat 

\else

\fi

\usepackage{hyperref}

\title{Can Authorship Attribution Models Distinguish \\ Speakers in Speech Transcripts?}

\author{
  Cristina Aggazzotti, Nicholas Andrews$^{*}$ 
  \\
  Johns Hopkins University
  \\
  USA
  \\
  \texttt{\{caggazz1, noa\}@jhu.edu}
  \And
  Elizabeth Allyn Smith$^*$ 
  \\
  Universit\'e du Qu\'ebec \`a Montr\'eal
  \\
  Canada
  \\
  \texttt{smith.elizabeth\_allyn@uqam.ca}
}

\date{}

\begin{document}
\maketitle
\begin{abstract}
Authorship verification is the task of determining if two distinct writing samples share the same author and is typically concerned with the attribution of \emph{written} text. In this paper, we explore the attribution of \emph{transcribed speech}, which poses novel challenges. The main challenge is that many stylistic features, such as punctuation and capitalization, are not informative in this setting. On the other hand, transcribed speech exhibits other patterns, such as filler words and backchannels (e.g., \emph{um}, \emph{uh-huh}), which may be characteristic of different speakers. We propose a new benchmark for speaker attribution focused on human-transcribed conversational speech transcripts. To limit spurious associations of speakers with topic, we employ both conversation prompts and speakers participating in the same conversation to construct verification trials of varying difficulties. We establish the state of the art on this new benchmark by comparing a suite of neural and non-neural baselines, finding that although written text attribution models achieve surprisingly good performance in certain settings, they perform markedly worse as conversational topic is increasingly controlled. We present analyses of the impact of transcription style on performance as well as the ability of fine-tuning on speech transcripts to improve performance.\footnote{Our benchmark is available at \href{https://github.com/caggazzotti/speech-attribution}{github.com/\\caggazzotti/speech-attribution}.}
\end{abstract}
\def\thefootnote{*}\footnotetext{Authors listed alphabetically.}
\def\thefootnote{\arabic{footnote}}

\section{Introduction}\label{sec:intro}\input{introduction}

\section{Related Work}\label{sec:related}\input{related}
\section{A Speaker Attribution Benchmark}\label{sec:benchmark}\input{benchmark}

\section{Experiments}\label{sec:experiments}\input{experiments}
\section{Discussion}\label{sec:discussion}\input{discussion}

\section*{Acknowledgments}\input{acknowledgments}

\bibliography{references}
\bibliographystyle{acl_natbib}

\appendix
\section{Appendix}\label{sec:appendix}\input{appendix}

\end{document}

%% file: introduction.tex
Identifying individuals based on their language use can be attempted using speech (speaker recognition) or writing (authorship attribution). Traditionally, speech data have been analyzed using phonetic features, such as pitch and articulation rate, or embeddings related to these features, such as x-vectors \citep{xvectors2018}, with broader linguistic information about the content of what is said playing little role \citep{gold-french2019, watt-brown2020}. Textual data, by contrast, have been analyzed historically via lexical, syntactic, semantic, and stylistic features \citep{mostellerwallace, stamatatos2018} and more recently via textual embeddings from neural networks \citep{ding2019, najafi-tavan2022}. 

There are a number of motivations for exploring the use of textual authorship attribution methods on transcribed speech. In challenging acoustic settings, such as with degraded audio or a disguised voice, speech content and syntax may be the only reliable signal available. In other cases, the original audio may no longer be available, leaving only a textual version of the speech. This is common with transcripts of interviews, court proceedings, and in commercial settings where text is preferred for archiving. Developing reliable identification of a speaker based on transcripts of their speech may also expose potential blind spots in current speaker anonymization approaches, which alter the speech signal but leave the speech content intact \citep{fang2019, sisman2021}.

Given ever-increasing spoken media (such as podcasts) and social communication as well as the growing popularity of automatic transcription systems (e.g.~\href{https://www.descript.com/}{Descript} for podcasters, \href{https://otter.ai/}{Otter} for meetings), there will be a need for understanding this domain and being able to accurately identify speakers, especially in forensic settings, in the future. Furthermore, understanding the generality of text attribution methods for transferring to not only new domains, but also new modalities (textual representations of auditory input) may indicate that these methods use deeper linguistic features rather than mainly relying on surface features, such as punctuation and capitalization. 

Despite the potential utility of distinguishing speakers via transcribed speech, there are a number of challenges. One that applies to both transcript-based and text-based attribution is the need for substantial writing samples to characterize style, which, in the speech domain, means a large number of utterances. Also, since transcribed speech is both a different domain and a different modality,\footnote{Although transcribed speech is in text form, it still originates from the modality of speech and is thus sufficiently different from other text domains.} text-based attribution models have more of a gulf to bridge than in a standard cross-domain task. For example, while written text contains features such as punctuation, spelling, and use of emoticons that can vary systematically across individuals, transcribed speech may lack these particular stylistic cues. Instead, speech contains other potentially identifying features, such as filler words (e.g.~\emph{um}), backchannels (e.g.~\emph{uh-huh}), and other discourse markers (e.g.~\emph{well}, \emph{I mean}) \citep{duncan1974, sacks1992}, and, more rarely, indications of pause length or intonation. Finally, working with transcribed speech, much like working with translated language, introduces noise into the system via the transcription or translation process. 

The present work contributes a benchmark for text-based authorship attribution models on human-transcribed conversational speech transcripts. Using the task of verification---i.e.\ determining if two transcripts have the same speaker or different speakers---we perform a systematic comparison of existing methods and provide a detailed analysis of the results. This work focuses on the following research questions. First, despite the difference in modality, do textual authorship models transfer to speech transcripts? Once we have established this benchmark, we probe other aspects of the task to help determine the limits of authorship models applied to transcripts. Since speech is transcribed based on a transcription style, what is the impact of transcription style on attribution performance? As alluded to above, many state-of-the-art authorship models focus on features such as capitalization that can be erased or standardized with transcription. Additionally, many models rely on topic as a clue for attribution, so to what extent does controlling for topic make the task harder? Next, does fine-tuning on speech transcripts significantly improve performance, and then, does further pre-training on speech transcripts improve it more? Finally, since in many forensic settings there are often limited data, how many utterances are required to achieve a given level of verification performance? By addressing these questions we provide a proof of concept for the viability of applying authorship models to speech transcripts.

%% file: related.tex
Early work in speaker identification for text used special-purpose models for novels that match utterances to the character who `said' them \citep{he2023-speakerID-novels}. The first work we are aware of to perform attribution on speech transcripts is a study that looks at word frequency to distinguish 250 speakers of Dutch \citep{scheijen2020}. In contrast, we consider a larger set of speakers and compare a range of different methods, both neural and non-neural. A contemporaneous related task is the PAN 2023 competition, which looked at cross-discourse type authorship verification between essays, emails, interviews, and speech transcripts \citep{pan2023}. They perform a weaker version of topic control, though, replacing named entities with generic tags. 
 
Recently, \citet{hansen2023} compare a number of statistical and neural authorship models on human spoken texts and large language models prompted to emulate spoken texts, finding that even simple $n$-gram-based authorship models can perform well on speech transcripts. However, we present contradictory results in this work, finding that most text-based authorship models have almost no predictive power once topic is controlled. We conduct further experiments that tease apart which factors influence performance, such as transcription style and number of utterances in the transcript.

The effect of conversational topic (and text genre) on text-based authorship attribution performance has been studied particularly in (forensic) stylometry to address cases of a domain mismatch between the texts of unknown authorship that are under investigation (test data) and the available comparison texts of known authorship (training data)~\citep{stamatatos2018}. For instance, an anonymous social media post might have to be compared to the news articles and blog posts of potential authors. These stylometric studies often focus on small amounts of data and/or few candidate authors, such as manually elicited data~\citep{baayen2002, goldstein2009} or select literary authors~\citep{kestemont2012}.  

Authorship studies on larger amounts of data across many authors use either no topic control or approximate topic control through domain labels, such as categories of Amazon product reviews~\citep{boenninghoff2019, zhu2021} and subreddits of Reddit~\citep{wegmann2022, zhu2021}. Although it is unclear how representative domain labels are of various topics, studies that implement some version of topic control generally find that performance decreases as topic control increases~\citep{wegmann2022}. 

%% file: benchmark.tex
To compare performance across the range of intended conditions, we focus on one dataset that fit all our requirements, namely large enough in number of speakers, number of utterances per speaker, and number of conversations to allow fine-tuning. We also focus on gold standard transcriptions since our aim is to establish an initial benchmark of performance under ideal conditions that can be used as a reference point for future experiments on noisier data. In fact, experimenting on the noisier output of automatic transcribers is out-of-scope for this paper but is an important next step in future work. To be able to adjust the difficulty level of the verification task to obtain a range of performance for the models, it was crucial that speakers discuss fixed topics, which would allow matching transcripts not only based on speaker, but also on the content of what is discussed. Finally, a conversational dataset aligns with many likely use cases of speaker attribution, especially in the forensic setting. 

We chose the Fisher English Training Speech Transcripts corpus \citep{fisher}, a collection of human-transcribed phone calls, due to its accessibility, size, manual error correction, conversational topic assignments, and gender balance amongst speakers. The corpus consists of 11,699 transcribed phone calls totalling 1,960 hours. Participants on calls generally did not know each other and had an assigned discussion topic that was randomly selected from a list. In general, participants stayed on topic throughout the duration of the call \citep{fisher}, which we confirmed through a manual check of a random sample of the transcripts. The calls lasted 10 minutes and speakers often participated in multiple calls.

The Fisher corpus contains two transcript ‘encodings’, or annotation styles. One was manually transcribed using WordWave quick transcription with error corrections and post-processing by BBN Technologies. This ‘BBN’ encoding includes prescriptive punctuation and capitalization according to an existing WordWave style guide \citep{kimball}. The Linguistic Data Consortium (LDC) provided the second encoding, with automatic segmentation of the audio data and manual transcription of the words, including a basic spell check \citep{fisher}. This transcription did not include punctuation (other than apostrophes and hyphens), put text in all lowercase, and often grouped together text by the same speaker despite interjected backchannels. Comparing performance on each encoding can help elucidate the extent to which the models capture ‘deeper’ authorship features rather than surface-level low-hanging fruit.~\autoref{fig:encodings} presents an example of each. Both encodings include non-speech sounds, such as laughing and undistinguished noise, in square brackets. The LDC encoding employs double parentheses for unclear productions that were guessed by the transcriber. 

\begin{figure}[hb!]\vspace{-.25cm}\small
{\bf BBN:} \vspace{-.2cm}
\begin{itemize}\setlength\itemsep{-.2em}
    \item[L:] Hi.  [LAUGH] So, do you have pets?
    \item[R:] Ah, no.
    \item[L:] Oh. I ha- --
    \item[R:] Do you?
    \item[L:] Yeah.  I do.  I have three dogs [LAUGH] --
    \item[R:] Oh, okay.
    \item[L:] -- and I have a bunch of fish.  I have --
    \item[R:] Oh.
    \item[L:] Yeah.  I have -- I have a black lab; he's eighty pounds, big guy. And then I have two little dogs, like terrier mixes [LAUGH].
\end{itemize}
{\bf LDC:}\vspace{-.2cm}
\begin{itemize}\setlength\itemsep{-.2em}
    \item[A:] hi [laughter] so do you have pets
    \item[B:] (( ah no )) 
    \item[A:] oh
    \item[A:] i ha- yeah i do i have three dogs [laughter]
    \item[B:] (( do you )) 
    \item[B:] oh okay
    \item[A:] and i have a bunch of fish i have yeah i have i have a black lab he's eighty pounds big guy and then i have two little dogs like terrier mixes
    \item[B:] (( oh ))
\end{itemize}\vspace{-.3cm}
\caption{\small Examples of the two Fisher transcript encodings, `BBN' and `LDC'.}
\label{fig:encodings}
\vspace{-9pt}
\end{figure}

We select the same transcripts from both encodings to test the impact of annotation on a model’s attribution performance, according to the following procedure. First, we use a 50/25/25 training/validation/test split with \textbf{no overlap in speakers} between training and evaluation splits, making the task more challenging. For the speakers in each set, verification trials are formed by matching two transcripts from either the same speaker (`positive') or different speakers (`negative'). Transcripts have~\url{~}1400 tokens across an average of 100 utterances. In~\autoref{sec:exp6} we vary the number of utterances used from each transcript; the results revealed that the appearance of names during introductions at the beginning of the call significantly helped model performance; as a result, we remove the first five utterances of each transcript for all experiments.\footnote{Many introductions concluded within the first two utterances per speaker but we removed five to be conservative. We also ran a simple check for (re)introductions later in the transcript but found them to be rare.}  

We create three different datasets according to level of difficulty by controlling (or not) for topic to the extent possible. The `base' level does not have any restrictions on topic: positive trials have one speaker on different calls while in negative trials, the speaker and call are different. The `hard' level introduces some topic control: positive trials consist of two transcripts from the same speaker in different calls in which the assigned discussion topic is different, and negative trials contain two transcripts from different speakers in calls in which the assigned topic is the same. The `harder' level contains the same positive trials as the `hard' level, but the negative trials are further restricted by only pairing speakers on the same call, so not only is the assigned topic the same, but the content within that topic also matches. In other words, the `hard' level is a rough measure of topic given that a number of subtopics can be discussed in ten-minute conversations, whereas the `harder' level is a more reliable measure of topic given that two people in the same call will cover the same range of subtopics. 

As a rough computational estimate of topic,~\autoref{table:trials} shows the percentage of noun (specifically, noun lemma) overlap between same speaker and different speaker transcripts. Although content can be conveyed through many parts of speech, \citet{wang2023} showed that masking only nouns is an effective way of obscuring content without obscuring (much) style. The rate of overlap across positive trials stays fairly consistent; however, the rate across negative trials increases with increased topic control, with over double the amount of overlap in the `harder' setting compared to the `base' setting, indicating that using the assigned conversation topics as a proxy for topic does have the intended effect to some extent. 

Running the authorship models on these datasets thus tests their ability to look beyond simplistic proxies for content and to utilize more structural features used by the speaker. For each difficulty level, the training set has \url{~}3300 verification trials, the validation set \url{~}1700, and the test set \url{~}1800 on average. The number of positive and negative verification trials per difficulty level is roughly balanced, cf.~\autoref{table:trials}.

\begin{table}
\scalebox{0.93}{
\setlength{\tabcolsep}{3pt}
\begin{tabular}{p{.45in}|cc|cc|c|c} 
  & \bf\%pos & \bf\%neg &\bf \#pos & \bf \#neg & \bf \#total & \bf \#spkrs\\
\toprule
\bf Base & 11.6 & 8.9 & 956 & 957 & 1913 & 1373\\ 
\bf Hard & 11.2 & 12.0 & 959 & 985 & 1944 & 1474\\
\bf Harder & 11.2 & 20.3 & 959 & 558 & 1517 & 1298 \\ \bottomrule
\end{tabular}}
\caption{\small Average \% of noun lemma overlap between transcripts in each positive/negative test set verification trial, \# of positive/negative/total test set trials per difficulty, and \# of speakers per difficulty level.}
\label{table:trials}
\vspace{-10pt}
\end{table}

%% file: experiments.tex
\textbf{Models}
We test and compare the performance of four main models. The first is Sentence-BERT (SBERT),\footnote{\href{https://huggingface.co/sentence-transformers/all-MiniLM-L12-v2}{huggingface.co/sentence-transformers/all-MiniLM-L12-v2}} a variant of the pretrained BERT network that creates semantically-related sentence embeddings \citep{sbert}. As a complement to the content-focused SBERT, we test Content-Independent Style Representations (CISR),\footnote{\href{https://huggingface.co/AnnaWegmann/Style-Embedding}{huggingface.co/AnnaWegmann/Style-Embedding}} which aims to capture writing style rather than content by controlling the topic of verification trials at training time \citep{wegmann2022}. The third model is an instance of Learning Universal Authorship Representations (LUAR),\footnote{\href{https://huggingface.co/rrivera1849/LUAR-MUD}{huggingface.co/rrivera1849/LUAR-MUD}} which does well with zero-shot transfer between Reddit, Amazon, and fanfiction stories \citep{rivera-soto2021}, capturing stylistic features of an author's writing with less content-sensitivity \citep{wang2023}. In summary, we use two models tailored to capture primarily content (SBERT) and primarily style (CISR), in addition to a model that combines both and does domain transfer well (LUAR). We specifically use a LUAR model trained on a large Reddit dataset consisting of comments by one million authors~\citep{khan2021}, which we hypothesize is more similar to the conversational speech transcripts than other more formal domains. 

The final model we consider is the recurrent neural network model AdHominem\footnote{\href{https://github.com/boenninghoff/AdHominem}{github.com/boenninghoff/AdHominem}} \citep{boenninghoff2019}, which performed well on speech transcripts in recent work \citep{hansen2023}. This model uses a hierarchical architecture to aggregate character, word, and sentence features from each document. For AdHominem, we converted the transcript trials into the appropriate format and then saved a model checkpoint, whose weights were restored for extracting features from the utterances in each trial. 

For reference, we also include two baselines: TF-IDF, a weighted measure of word overlap, and PAN's authorship verification baseline of TF-IDF-weighted character 4-grams (PANgrams)\footnote{\href{https://github.com/pan-webis-de/pan-code/tree/master/clef23/authorship-verification}{github.com/pan-webis-de/pan-code/tree/master/clef23/authorship-verification}} \citep{pan2023}.\footnote{We did not include the best performers (all SBERT-based models) on the contemporaneous PAN 2023 authorship verification competition, whose task involved written data and transcribed spoken data, because neither the models nor the training data are publicly available.} 

In line with the research questions in~\autoref{sec:intro}, we set up the experiments as follows.~\autoref{sec:exp1} creates a performance benchmark of the aforementioned models evaluated \emph{out-of-the-box} on speech transcripts.~\autoref{sec:exp2} tests these out-of-the-box models on both the BBN and LDC encodings to determine the effect of transcription style on performance. This includes an additional comparison between the default LUAR pre-trained on Reddit and an instantiation of LUAR pre-trained on Reddit that has been normalized to look like the LDC data.~\autoref{sec:exp3} varies the difficulty level of the task (`base', `hard', `harder') by controlling for the topic discussed across the verification trials.~\autoref{sec:exp4} adds a step of fine-tuning on each training set difficulty level and then evaluates the models on each difficulty level of the test set. As a first look into the impact of pre-training domain on performance,~\autoref{sec:exp5} tests the best performing model, LUAR, pre-trained on the speech transcripts themselves. Finally,~\autoref{sec:exp6} varies the number of utterances used in each transcript from 25 to the full transcript.

\subsection{Experiment 1: Baselines}
\label{sec:exp1}

\begin{table*}[t]
\centering
\resizebox{\linewidth}{!}{%
\small
\begin{tabular}{lccc|ccc|ccc}
 & \multicolumn{3}{c}{\emph{BBN encoding}} & \multicolumn{3}{c}{\emph{LDC encoding}} & \multicolumn{3}{c}{\emph{NormLDC encoding}}\\
\toprule
\bf Model & \bf Base & \bf Hard & \bf Harder & \bf Base & \bf Hard & \bf Harder & \bf Base & \bf Hard & \bf Harder \\ \midrule
\bf TF-IDF$_o$ & \underline{0.758} & \underline{0.676} & 0.514 & 0.759 & 0.680 & 0.519 & \underline{0.790} & \underline{0.705} & \underline{0.546} \\ 
\bf PANgrams$_o$ & \bf0.796 & \bf0.699 & \bf0.547 & \bf0.805 & \underline{0.710} & \underline{0.572} & \underline{0.790} & 0.691 & \bf0.564  \\
\bf AdHominem$_o$ & 0.565 & 0.553 & 0.492 & 0.600 & 0.574 & 0.536 & 0.588 & 0.589 & \underline{0.545}  \\
\bf SBERT$_o$ & 0.646 & 0.483 & 0.322 & 0.653 & 0.456 & 0.283 & 0.621 & 0.514 & 0.174 \\
\bf CISR$_o$ & 0.612 & 0.578 & \underline{0.534} & 0.680 & 0.664 & \bf0.646 & 0.622 & 0.532 & 0.493 \\ 
\bf LUAR$_o$ & 0.714 & 0.633 & 0.472 & \underline{0.803} & \bf0.711 & 0.547 & \bf0.837 & \bf0.722 & \underline{0.543} \\
\bottomrule
\end{tabular}%
}
\caption{\small Bootstrapped test performance (AUC) across all out-of-the-box ($_o$) models for the BBN, LDC, and NormLDC encodings across all difficulty levels. Best performance for each encoding and difficulty level is bolded and second best, underlined, the differences of which are all statistically significant ($p<0.002$) except for ties. The largest standard error is 0.0004.}
\label{table:box}
\vspace{-9pt}
\end{table*}

To address whether text-based authorship models transfer to transcribed speech, we evaluate the baselines (TF-IDF$_o$, PANgrams$_o$) and main models (AdHominem$_o$, SBERT$_o$, CISR$_o$, LUAR$_o$) out-of-the-box on the `base' difficulty level verification trials. Recall that each transcript originally had \url{~}1400 tokens and \url{~}100 utterances on average, but we removed the first five utterances per speaker as we confirmed they contain name and topic introductions in most transcripts.

For the TF-IDF baseline, the vectorizer was fit to the Reuters-21578 corpus, which contains 10,788 news documents and totals 1.3 million words \citep{reuters},\footnote{We also tried fitting the vectorizer to the training set transcripts but found worse results so fit to Reuters to focus on (better) out-of-the-box performance of written text models.} the document-term matrix obtained for each transcript in the test set trials, and the cosine similarity calculated between each matrix in each trial. For the out-of-the-box PAN baseline, PANgrams$_o$ was trained on the most recent openly available PAN authorship verification dataset, fanfiction stories \citep{PAN2020}, and evaluated on the test set trials using cosine similarity. 

For the main models, we first obtain an embedding for each transcript in the trial,\footnote{While LUAR is hierarchical, accepting \emph{multiple} utterances as independent inputs, SBERT and CISR encode representations on the individual sentence (or document) level. To accommodate multiple inputs with SBERT and CISR, we embed each utterance independently and obtain the final embedding using the coordinate-wise mean of all utterance vectors. We found that this typically worked better than concatenating the text of all utterances and producing a single embedding.} then calculate the cosine similarity between each set of two embeddings. AdHominem is trained using Euclidean distance, so we negate it to compute speaker similarity between pairs of transcripts. 

We evaluate the performance of all models by bootstrapping the area under the receiver operating characteristic curve (AUC) score with 1000 resamples of the test data. For testing statistical significance between the first and second best performers, as indicated in all tables in the paper, we used a paired t-test over 1000 resamples to test the null hypothesis that the AUC (or EER) scores produced by two models are the same.\footnote{We also ran a non-parametric test, the Wilcoxon signed-rank test, but since the results were similar, we report the (more conservative) results from the more powerful parametric paired t-test.}

\paragraph{Out-of-the-box attribution models transfer to speech transcripts (without topic control).} 
AUC score test set results are in~\autoref{table:box} and EER results are in~\autoref{table:eer} in~\autoref{sec:appendix}. Focusing on the leftmost column (BBN `text-like' encoding, `base' difficulty), there is a clear ranking in performance: PANgrams$_o$ achieves the highest performance followed by TF-IDF$_o$, LUAR$_o$, SBERT$_o$, and CISR$_o$, with all scores well above chance (AUC = $0.5$). AdHominem$_o$ performs the worst but still above chance. These initial results---considering only the `base' setting for now---suggest at least some transfer from the text domain to the speech domain. However, we revisit this idea in~\autoref{sec:exp3} when discussing results on topic-controlled datasets.

\subsection{Experiment 2: Transcription style}
\label{sec:exp2}

To test the impact of transcription style, we ran~\hyperref[sec:exp1]{Experiment 1} (with the same verification trials) on both the BBN encoding, with punctuation and capitalization, and the LDC encoding, with limited or none. Overall, we find that transcription style can have a surprisingly large impact on performance.

\paragraph{Superficial text features are not needed.} Comparing the leftmost column (`base') of the left (BBN) and middle (LDC) sections of~\autoref{table:box}, we see that TF-IDF$_o$, PANgrams$_o$, and SBERT$_o$ are similar for both encodings. A lack of difference for these models is expected since semantic content is similar across embeddings. AdHominem$_{o}$ shows some improvement from BBN to LDC, but there are significant increases for CISR$_o$ and especially LUAR$_o$ with the LDC encoding. The ability of LUAR$_o$ to perform well out-of-the-box, and to perform significantly better on the normalized transcription style in general, suggests that LUAR does not rely on `superficial' prescriptive textual features. The fact that model performance does not degrade on normalized data, even improving in some cases, is a promising sign for potential applications to \emph{automatically} transcribed speech, which often lacks such features. 

\begin{table}[b]\vspace{-.3cm}
\centering
\begin{tabular}{lccc}
\multicolumn{4}{c}{\emph{Base} level normalization variations: AUC}\\
\toprule
\bf Model & \bf BBN & \bf LDC & \bf NormLDC \\ \midrule
\bf LUAR$_o$ & 0.714 & 0.803 & \bf0.837\\
\bf LUARnorm$_o$ & 0.717 & 0.794 & 0.831\\  
\bottomrule
\end{tabular}
\caption{\small Bootstrapped test performance (AUC) across out-of-the-box ($_o$) LUAR and normalized LUAR models for the BBN, LDC, and NormLDC encodings at the \emph{base} difficulty level. Best performance overall is bolded. The largest standard error is 0.0003 and all differences are statistically significant ($p<0.001$).}
\label{table:luar}
\vspace{-9pt}
\end{table}

\paragraph{Removing transcript annotations can improve performance.} Since LUAR showed the biggest difference between encodings, we ran a LUAR model pre-trained on the same subset of the Reddit dataset as before, this time normalized to look like the LDC encoding. The normalization included lowercasing and removing all HTML special entities, hyperlinks, emoticons, and punctuation except apostrophes and hyphens between letters. This model is called \textbf{LUARnorm} and its out-of-the-box performance on the `base' level is in~\autoref{table:luar}. LUAR$_o$ and LUARnorm$_o$ achieve similar performance. (The performance of both models across all encodings and difficulties is shown in~\autoref{table:luarall} (AUC) and~\autoref{table:eerluar} (EER).) 

Since the speech transcripts contain bracketed non-speech sounds and annotators' hypothesized transcriptions, the normalized Reddit dataset and LDC encoding are still not exactly equivalent. Thus, we also further normalized the LDC encoding data, creating a \textbf{NormLDC} encoding by removing the brackets and double parentheses. This new encoding now more closely resembles the normalized Reddit, leaving only differences between text and speech characteristics. Both LUAR$_o$ and LUARnorm$_o$ perform best on the NormLDC encoding out of all three encodings and continue to perform roughly similarly. These results suggest that a lack of capitalization and punctuation (LDC), along with the removal of transcript-specific annotations (NormLDC), seem most influential for improving performance, but pre-training on normalized text data (LUARnorm) does not significantly impact performance.

\subsection{Experiment 3: Topic}
\label{sec:exp3}

To test the effect of topic, after running~\hyperref[sec:exp1]{Experiment 1} on the `base' dataset, we ran it on the `hard' and `harder' datasets. Since performance of both LUAR$_o$ and LUARnorm$_o$ was highest on the NormLDC encoding previously, we included it here as well. The full AUC results across all settings are shown in~\autoref{table:box} and the corresponding EER results are in~\autoref{table:eer}. 

\paragraph{Out-of-the-box performance degrades with topic control.} Across all three encodings, the `hard' dataset had lower AUC scores than the `base' dataset, and the `harder' dataset lower than the `hard' dataset. In particular, TF-IDF$_o$ and AdHominem$_o$ decrease consistently to around chance. Even PANgrams$_o$, one of the best performers on the `base' level, degrades significantly, suggesting that simple $n$-gram approaches are incapable of capturing speaker stylistic features and, as a result, fail under topic shift. SBERT$_o$ was the most severely affected by the topic manipulation, with performance on the hardest dataset well below chance. As we expect SBERT$_o$ to exploit content differences that we successively remove with this manipulation, this result is not surprising. CISR$_o$ also had a decrease in performance but to a smaller extent, likely because its training involves specific attempts at controlling for topic by using subreddits as proxies for topic in the Reddit training data~\citep{wegmann2022}.

For LUAR$_o$, we see \url{~}10\% performance loss between the `base' and `hard' conditions, and \url{~}15\% between the `hard' and `harder' conditions, to near chance. We propose two potential explanations for the greater difference between the `hard' and `harder' conditions. First, the `hard' condition does not fully accomplish its topic manipulation as proxies for topic diverge from linguistic definitions of thematic topic, discourse topic, and the like. As two examples, `summer plans' may include discussions of both vacations and temp jobs, with little semantic overlap between these topics, and a range of subtopics may be discussed throughout each conversation. With the `harder' dataset, in which each speaker in the trial discusses largely the same topic(s) and subtopic(s) in the same order given that they represent each side of the same conversation, we expect a higher degree of topic identity. 

A second possibility comes from the literature on linguistic accommodation. Peers in conversation adapt their speech style to more closely resemble that of their interlocutor \citep{danescu,pardo2022,giles2023}. There are a number of reasons, both automatic and intentional, for this kind of convergence, but for our purposes, all factors relevant to the Fisher corpus would favor it (i.e.~speakers unknown to each other in a collaborative task wanting to make a favorable impression). If our speakers accommodated to one another, their styles of speaking would become more similar the longer they talked and therefore more difficult to distinguish. This would explain the supplemental difficulty we find with the `harder' dataset. These explanations are independent and could both contribute; future work will quantify the effect of accommodation, similar to what \citet{danescu} did for Twitter exchanges, in order to tease these apart.

Overall, since topic shifts are expected to some degree in many applications of speaker verification, these results suggest that methods developed for written text are inadequate out-of-the-box for verification of transcribed speech. CISR$_o$ is a promising approach for being more resilient to topic control, but performs worse compared to other models on the `base' and `hard' levels.

\subsection{Experiment 4: Fine-tuning}
\label{sec:exp4}

\begin{table*}[t]
\centering
\resizebox{\linewidth}{!}{%
\small
\begin{tabular}{lccc|ccc|ccc}
\multicolumn{10}{c}{\emph{LDC Encoding}}\\ \toprule
\emph{Trained on:} & \multicolumn{3}{c}{\bf{Base}} & \multicolumn{3}{c}{\bf{Hard}} & \multicolumn{3}{c}{\bf{Harder}}\\
\midrule
\emph{Evaluated on:} & \bf Base & \bf Hard & \bf Harder & \bf Base & \bf Hard & \bf Harder & \bf Base & \bf Hard & \bf Harder \\ \toprule
\bf TF-IDF$_{ft}$ & 0.535 & 0.500 & 0.502 & 0.506 & 0.593 & 0.537 & 0.513 & 0.569 & 0.533 \\
\bf PANgrams$_{ft}$ & \underline{0.763} & 0.628 & 0.419 & \underline{0.764} & 0.623 & 0.412 & \bf0.765 & 0.628 & 0.417 \\
\bf AdHominem$_{ft}$ & 0.586 & 0.578 & 0.541 & 0.594 & 0.576 & 0.546 & 0.542 & 0.558 & 0.586 \\
\bf SBERT$_{ft}$ & 0.694 & 0.650 & 0.632 & 0.589 & \underline{0.830} & \bf0.835 & 0.530 & \underline{0.818} & \bf0.935 \\
\bf CISR$_{ft}$ & 0.722 & \underline{0.696} & \underline{0.744} & 0.660 & 0.642 & 0.690 & 0.674 & 0.651 & 0.781 \\
\bf LUAR$_{ft}$ & \bf0.844 & \bf0.818 & \bf0.753 & \bf0.798 & \bf0.872 & \underline{0.818} & \underline{0.694} & \bf0.820 & \underline{0.894} \\\bottomrule
\end{tabular}%
}
\caption{\small Bootstrapped test performance (AUC) across all fine-tuned ($_{ft}$) models for the LDC encoding across all distribution combinations. Best performance per combination is bolded and second best, underlined, the differences of which are all statistically significant ($p<0.001$). The largest standard error is 0.0004.}
\label{table:ft}
\vspace{-9pt}
\end{table*}

To see whether fine-tuning on speech transcripts can improve transfer from authorship models to speech transcripts, we fit a multilayer perceptron (MLP) classifier to the concatenated embeddings, from each model, of each trial in the training set verification trials.\footnote{Model selection was performed based on validation performance. We found that other fine-tuning approaches, such as linear models, performed worse on validation data. Hyperparameter optimization experiments found that using the Adam solver with 800 maximum iterations worked best overall across models on the validation trials. We kept all other default values and used a random state of 1.} Since the available number of trials is limited based on the size of the corpus, we are wary of overfitting the data and thus fit a transformation of the (fixed) embedding from each model rather than fine-tune the whole model. We then evaluate the classifier on its probability predictions of the concatenated embeddings of the test trials. To account for variation, we bootstrap the AUC score over 1000 random resamples of the test trials. We use this procedure for all models except PANgrams$_{ft}$, which directly uses the PAN-provided code to train and calculate cosine similarity within trials~\citep{pan2023}, but we additionally bootstrap its AUC score over 1000 random test trial resamples to produce a more robust evaluation. 

The classifier is fine-tuned on the training set for each difficulty level and then evaluated on all test sets, e.g.~train on `base' and evaluate this model on `base', `hard', and `harder'.~\autoref{table:ft} gives these results for the LDC encoding, which produced higher scores overall than the BBN encoding.~\autoref{table:ftbbn} shows the results for the BBN encoding; EER results for the train-test match setting across the BBN, LDC, and NormLDC encodings are provided in~\autoref{table:eer}. 

\paragraph{Fine-tuning helps if train-test distributions match.} For most models, training on the same difficulty level as that of the evaluation data yields the best performance. This is especially true in the harder settings, with each model's highest performance across all settings achieved in the `harder'-`harder' train-test distribution setting. Since the models are evaluated on trials of the same difficulty as they are trained on, it makes sense that fine-tuning the neural models on difficult trials does indeed help with harder cases. 

The SBERT$_{ft}$ model achieves the best performance in the `harder'-`harder' condition (0.935). Unlike SBERT$_o$, which was not able to overcome the reduction in content differences in the `harder' setting, SBERT$_{ft}$ seems to have `learned' to take advantage of the content similarity. Returning to rates of noun overlap in~\autoref{table:trials}, there is almost double the noun overlap between transcripts of different speakers in the same conversation (negative trials in the `harder' setting) than between transcripts of the same speaker in conversations on different topics (positive trials in the `harder' setting). One hypothesis for SBERT$_{ft}$'s higher performance, then, is that it used the rate of noun overlap as a \emph{negative} indicator of whether the speaker is the same or not, with higher noun overlap rates indicating different speakers and lower rates indicating the same speaker. In other words, SBERT$_{ft}$ used a shortcut that may have worked well in this experimental setup, but likely would not work well in other setups, which is corroborated by its poorer performance on the `harder'-`base' and `harder'-`hard' settings. 

LUAR$_{ft}$ has the next best performance after SBERT$_{ft}$, also in the `harder'-`harder' condition, and shows a strong correspondence between distribution matching. CISR$_{ft}$, though, does not show this same pattern. Again, as its training attempts to control for topic, its performance is less affected by our topic manipulation, and fine-tuning on particular settings does not show as significant of a difference. AdHominem$_{ft}$ could be similar, though its performance in general is much lower. 

Unlike the out-of-the-box models, the fine-tuned models can, to an extent, overcome challenges, such as less-than-perfect topic control and speaker style accommodation; exposure to trials of the same, or similar, difficulty level in training enables them to encode identifying stylistic features of speakers beyond the conversation topic. However, there is no general-purpose model that works well across difficulty levels; models work best when they are trained and evaluated on data from the same distribution. These models have potential for further improvement with more specialized tuning, which is a direction for future work.

\subsection{Experiment 5: Pre-training domain}
\label{sec:exp5}

To address how pre-training domain for style representation impacts performance, we tried pre-training specifically on speech transcripts. Since LUAR was the overall best performer in the previous experiments, we conducted focused experiments on LUAR to test this question. 

\begin{table}[t]
\centering
\begin{tabular}{lccc}
 & \multicolumn{3}{c}{\emph{NormLDC encoding}} \\
\toprule
\bf Model & \bf Base & \bf Hard & \bf Harder \\ \midrule
\bf LUAR$_o$ & 0.837 & 0.722 & 0.543 \\
\bf LUARnorm$_o$ & 0.831 & 0.704 & 0.524 \\  
\bf PreTrain-BBN$_o$ & 0.887 & 0.709 & 0.505 \\ 
\bf PreTrain-LDC$_o$ & \underline{0.906} & 0.699 & 0.418 \\\midrule
\bf LUAR$_{ft}$ & 0.844 & 0.869 & 0.876 \\
\bf LUARnorm$_{ft}$ & 0.864 & 0.875 & 0.907 \\
\bf PreTrain-BBN$_{ft}$ & \underline{0.906} & \underline{0.909} & \bf0.952 \\ 
\bf PreTrain-LDC$_{ft}$ & \bf0.909 & \bf0.921 & \underline{0.935} \\\bottomrule
\end{tabular}%
\caption{\small Bootstrapped test performance (AUC) across $_o$ and $_{ft}$ LUAR models for the NormLDC encoding across all train-test matched difficulty levels. Differences between first and second best are all statistically significant ($p<0.001$) except between ties. The largest standard error is 0.0004.}
\label{table:pretrain}
\vspace{-10pt}
\end{table}

\paragraph{Pre-training on speech transcripts performs best.} PreTrain-BBN and PreTrain-LDC are two separate instantiations of LUAR that were pre-trained on the full training set of speech transcripts, without refining by difficulty level, for the BBN and LDC encoding, respectively. Out-of-the-box, these models followed the previously described pipeline of being evaluated using cosine similarity on the validation (and later test) set. In the fine-tuned case, again following the same protocol as before, an MLP classifier was trained on the training set verification trials (i.e.~the same training data as seen in pre-training, but this time as trials of a particular difficulty) and evaluated using bootstrapped AUC score on the validation (and later test) set. Since training and evaluating on the same difficulty performed best in~\hyperref[sec:exp4]{Experiment 4}, we restrict the experiment to the train-test distribution match condition across difficulty levels. Performance was best on the NormLDC encoding, so~\autoref{table:pretrain} focuses on these results, but the results for all encodings are shown in~\autoref{table:luarall} (AUC) and~\autoref{table:eerluar} (EER). 

As expected, compared to the other LUAR models, PreTrain-BBN$_{ft}$ and PreTrain-LDC$_{ft}$ perform best across all three difficulties and achieve the highest performance of any model on the `harder' level. On the `base' level, PreTrain-BBN$_o$ and PreTrain-LDC$_o$ are fairly close seconds to their fine-tuned counterparts; however, in the `hard' and `harder' levels, PreTrain-BBN$_o$'s and PreTrain-LDC$_o$'s performance decreases significantly. This drop-off is most likely due to the train-test mismatch between pre-training on all training transcripts and evaluating only on verification trials of a particular difficulty level. The fine-tuned models suggest, though, through their increased performance across levels, that during pre-training, LUAR encodes speech-specific features that the MLP can avail of.

\subsection{Experiment 6: Varying input size}
\label{sec:exp6}

\begin{figure}[ht!]
    \includegraphics[width=.48\textwidth]{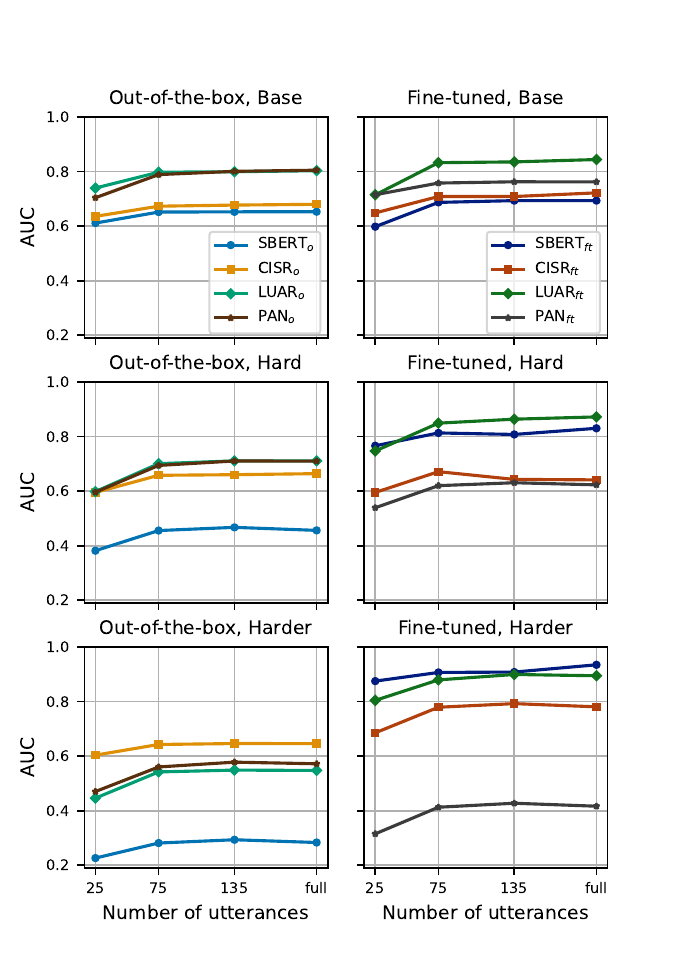}
\caption{\small Bootstrapped AUC test performance (y-axis) across out-of-the-box and fine-tuned models (columns) on the LDC encoding at the 3 levels of difficulty (rows) with the number of utterances per speaker varied (x-axis). Increasing the number of utterances improves performance for all models, with the best generally achieved by 135 utterances.}
\label{fig:utts}
\vspace{-9pt}
\end{figure}

Our final experiment tested the impact that observing more data has on attribution by varying the number of utterances used in each verification pair. We ran~\hyperref[sec:exp1]{Experiment 1} and~\hyperref[sec:exp4]{Experiment 4} (with training and evaluation distributions matched) on trials of transcripts containing incrementally more utterances, ranging from the first 25 per speaker to the full transcript. Speakers averaged \url{~}95 utterances per transcript (after the first 5 were removed) and the longest had \url{~}200 utterances. We chose the four best performing models for this experiment but only one LUAR (the standard instantiation for better comparability): PANgrams, SBERT, CISR, and LUAR. Since performance across models was not consistently better on the NormLDC encoding, we ran this experiment only on the BBN and LDC encodings. The graphs for the LDC encoding are in~\autoref{fig:utts} and those for the BBN encoding are in~\autoref{fig:uttsbbn}. These display the AUC score performance for each out-of-the-box (left column) and fine-tuned model (right column) on transcripts of length 25, 75, 135, and the full number of utterances per speaker in each pair. Each row represents an increase in difficulty level. 

\paragraph{Performance plateaus after 75 utterances.} Across both out-of-the-box and fine-tuned models, all increase in performance from 25 to 75 utterances, begin to plateau after 75, and some reach slightly higher performance by 135 utterances. Only 22\% of the LDC test transcripts are less than 75 utterances long and 67\% are less than 100 utterances, so most transcripts have not concluded at 75. 

\begin{table*}[t]
\centering
\begin{tabular}{lcc|cc|cc}
 & \multicolumn{6}{c}{\emph{LDC Encoding}} \\\toprule
 & \multicolumn{2}{c}{\textbf{Base}} & \multicolumn{2}{c}{\textbf{Hard}} & \multicolumn{2}{c}{\textbf{Harder}}\\
\midrule
\bf Model & \bf first 50 & \bf last 50 & \bf first 50 & \bf last 50 & \bf first 50 & \bf last 50 \\ \toprule
\bf PANgrams$_o$ & \underline{0.741} & \underline{0.757} & \underline{0.620} & \underline{0.683} & 0.461 & \underline{0.519} \\
\bf SBERT$_o$ & 0.648 & 0.675 & 0.467 & 0.654 & 0.244 & 0.309 \\
\bf CISR$_o$ & \bf0.747 & 0.646 & \bf0.682 & 0.664 & \bf0.637 & \bf0.600 \\  
\bf LUAR$_o$ & 0.723 & \bf0.782 & \bf0.682 & \bf0.767 & \underline{0.495} & 0.441 \\ \bottomrule
\end{tabular}%
\caption{\small Bootstrapped test performance (AUC) across out-of-the-box ($_o$) models for the LDC encoding across all difficulty levels for trials restricted to each speaker's first 50 utterances or last 50 utterances. Best performance for each difficulty level and transcript section is bolded and second best, underlined, the differences of which are all statistically significant ($p<0.001$) except for ties. The largest standard error is 0.006.}
\label{table:chunks}
\vspace{-10pt}
\end{table*}

To help determine if different parts of the transcript contribute more speaker information than others, we also ran a preliminary experiment attributing transcripts using only the beginning or the end of the transcript. Specifically, we restrict to transcripts having at least 100 utterances and create two further evaluation datasets. The first is similar to our existing experiments but restricted to the first 50 utterances to produce the speaker representation. The second instead takes the last 50 utterances to produce the speaker representation. Then we construct three experiments for our `base', `hard', and `harder' settings using the same protocol as elsewhere in the paper. Since there were significantly fewer trials (\url{~}100 per difficulty) after requiring each speaker per trial to have at least 100 utterances, these results should be compared relatively to each other, but not to other results in the paper.~\autoref{table:chunks} reports these results for the same four best performing out-of-the-box models on the LDC encoding. 

We make the following observations. First, we see that the baseline model PANgrams$_o$ and the semantic model SBERT$_o$ consistently benefit from using the last 50 utterances across all difficulty levels. LUAR$_o$, which performs better in the `base' and `hard' settings, also benefits from using the last 50 utterances over the first 50, except in the `harder' setting. The CISR$_o$ style representation, however, exhibits the opposite trend as PANgrams$_o$ and SBERT$_o$, being consistently worse when using the last 50 utterances. Overall, further study is needed to understand these differences in performance, although we may hypothesize that accommodation, which is likely to be evident later in a conversation, is playing a role in explaining these results, particularly for CISR$_o$. 

%% file: discussion.tex
\paragraph{Summary of findings} 
The primary goal of this work was to provide a proof of concept and establish baseline performance of text-based authorship models on speech transcripts. We focused on gold standard, human-transcribed transcripts as a starting point to find an upper bound on performance since this task has not previously been benchmarked. In so doing, we discovered the following answers to our research questions. First, despite the modality difference, we found that off-the-shelf textual authorship models, such as PANgrams and LUAR, transfer surprisingly well to speech transcripts, unless we control for topic, in which case all models' performance drops drastically. This finding contradicts previous work that did not rigorously control for topic, suggesting that model performance may have resulted from spurious correlations with topic rather than an ability to distinguish speakers. 

Transcription style also impacts performance, with most models performing best on the LDC transcripts normalized to remove capitalization and punctuation. Further normalization to remove speech transcript-specific annotations, such as brackets around non-speech sounds and double parentheses around annotators’ hypothesized transcriptions (NormLDC), hurt performance for many models, however. Adding superficial prescriptive textual features to transcripts is thus perhaps an unneeded processing step, but maintaining a distinction between annotations and regular speech is. 

We found that fine-tuning on speech transcripts significantly improves performance for most neural models, with the best performance achieved when the training and evaluation data are drawn from the same difficulty-level distributions, specifically in the `harder' condition where negative pairs are drawn from the same conversation. The `base' setting, though, represents a complete sample of pairwise verification trials without any artificial subsampling; therefore, while it is our easiest setting, it also represents a possibly more realistic setting than `hard' and `harder' cases. Choice of model should thus be determined by the particularities of the available data and the specific application. Separately, we find that additionally pre-training the model on speech transcripts can further improve performance. Finally, performance across all models plateaus after 75 utterances, despite most transcripts containing at least 20 more utterances, but which section (beginning or end) of each call is most useful for speaker attribution differs by model.

\paragraph{Limitations and future work}
Our best results use fine-tuned author representations pre-trained on the same speech transcripts; future work should explore variations that might produce even better performance, such as pre-training a dedicated model on a larger and more diverse dataset of speech transcripts. To better understand how the models are performing, future work should conduct a qualitative analysis of the results, linguistically examining which trials the models predict correctly and incorrectly to find any consistencies across models as well as any features the models might be using to make their determinations.

We note that in our `harder' dataset, accommodation may play a role in the results in addition to topic control, but do not tease apart the relative impact of this and other factors. Future work should attempt to quantify the amount of accommodation that occurs between speakers in the same call, which could also further inform which stages of conversation are most revealing of speaker style. An eventual extension might also look at the extent to which topics change over the course of conversations with specified discussion subjects, though a qualitative evaluation of some of the corpus indicated less evolution than expected.

Adding more baselines, such as a linguistic stylometric method, which calculates the frequencies of features at various linguistic levels, such as part-of-speech tags and function words, can also provide more informative comparisons. To get a better sense of how generalizable these results are to other speech domains, future work should include non-conversational data (e.g.~speeches) and other conversational forms (e.g.~interviews). The range of experiments should also be run on different languages (Fisher has Spanish and Arabic corpora) for direct comparison with the results of this work. 

Finally, for many real world applications, manually annotating or correcting transcripts to produce gold standard transcriptions is unfeasible. Transcripts will thus have varying amounts of noise, impacting attribution performance. Future work should investigate this question by running the same speech samples through several automatic transcribers, measuring the amount of noise and comparing model performance across these noisier transcripts.

\paragraph{Broader impact} 
As previously mentioned, analyzing the content and style of what is said in addition to the speech signal itself could improve speaker recognition performance, especially in low-quality acoustic settings (and can be the only option with discarded audio, etc.). Forensic settings, in particular, often have very little and/or degraded audio data, so combining insights across all linguistic levels may enrich current models and provide a more comprehensive speaker profile. Even in cases with good quality audio, having an independent method reach the same conclusion would help confirm speaker recognition results, providing more confidence in the attribution. 

A related observation is that speaker anonymization methods currently tend to obfuscate the acoustic signal, leaving the speech content, syntax, etc. intact. Since our results showed that even out-of-the-box models can perform well on verifying speakers based strictly on the remaining linguistic features of their speech as transcribed, this finding exposes a current weakness of speaker anonymization models that should be addressed in order to more comprehensively protect speakers' identities. In settings for which we imposed a stringent control for topic, though, attribution performance dropped considerably, suggesting that textual attribution models do need to be adapted to the speech domain in order to more robustly attribute speakers based on their transcribed speaking style. Nonetheless, models like CISR, which appear to be more robust against topic control out of the box, suggest that there is some overlap between topic-independent writing style and transcribed speaker style.

Finally, testing authorship models on the new domain of speech transcripts provides further insights into how the models work, especially `black box' neural models. Through these experiments, we obtain a better understanding of not only the abilities and limitations of authorship models, allowing us to apply them more accurately and effectively, but also the similarities and differences between written and spoken data. 

\paragraph{Ethical considerations}
Our findings should be carefully interpreted before considering speaker attribution for any real-world applications. For instance, although fine-tuning a model in our `harder' setting can significantly improve its performance on the same `harder' condition, this is an artificial setting that is not generally representative of real-world distributions of speaker transcripts. In addition, we find that performance across all models significantly decreases with a more rigorous control for topic, indicating that it would be premature to apply these models if topic shifts or topic differences across samples are possible in the application domain. We acknowledge that further enhancements of the methods presented, such as a better accounting of topic, may be used to defeat speaker anonymization systems, but our results suggest that current methods are not yet robust enough to topic manipulations to have this capability. Regardless of topic, though, these models should not be applied in domains in which it is important to understand how and why an attribution decision is made; such models would not pass the Daubert standard~\citep{daubert} for scientific evidence in the U.S. judicial context, for example.  

Many of the models we used were trained on anonymized social media data that have implicit biases, such as imbalances in the prevalence of authors from certain demographic groups. As a result, attribution performance may vary based on the same latent demographic factors, which is an issue that needs further study. One remediation may be to control for demographics when training the neural representations and then ensure that within-group performance is consistent for all relevant factors. Finally, it is worth noting that population-level statistics do not currently exist to determine the extent to which people speak (or write) like one another---while it is tempting to think that we have unique patterns of speech based on the success of some attribution models, we still have no real understanding of how rare certain styles are.

%% file: acknowledgments.tex
We thank the TACL reviewers and action editor for their insightful comments. We also thank Rafael Rivera Soto for his advice on model fine-tuning and help with model pre-training.

This research is supported in part by the Office of the Director of National Intelligence (ODNI), Intelligence Advanced Research Projects Activity (IARPA), via the HIATUS Program contract \#D2022-2205150003. The views and conclusions contained herein are those of the authors and should not be interpreted as necessarily representing the official policies, either expressed or implied, of ODNI, IARPA, or the U.S. Government. The U.S. Government is authorized to reproduce and distribute reprints for governmental purposes notwithstanding any copyright annotation therein.

%% file: appendix.tex
\begin{table*}[!htb]
\centering
\resizebox{\linewidth}{!}{%
\small
\begin{tabular}{lccc|ccc|ccc}
\emph{\textbf{AUC}} & \multicolumn{9}{c}{\emph{BBN Encoding}}\\ \toprule
\emph{Trained on:} & \multicolumn{3}{c}{\bf{Base}} & \multicolumn{3}{c}{\bf{Hard}} & \multicolumn{3}{c}{\bf{Harder}}\\
\midrule
\emph{Evaluated on:} & \bf Base & \bf Hard & \bf Harder & \bf Base & \bf Hard & \bf Harder & \bf Base & \bf Hard & \bf Harder \\ \toprule
\bf TF-IDF$_{ft}$ & 0.536 & 0.506 & 0.498 & 0.508 & 0.595 & 0.538 & 0.514 & 0.567 & 0.531 \\
\bf PANgrams$_{ft}$ & \underline{0.756} & 0.637 & 0.419 & \bf0.759 & 0.633 & 0.413 & \bf0.759 & 0.640 & 0.419 \\
\bf AdHominem$_{ft}$ & 0.582 & 0.554 & 0.513 & 0.565 & 0.557 & 0.520 & 0.518 & 0.524 & 0.595 \\
\bf SBERT$_{ft}$ & 0.689 & \underline{0.640} & \underline{0.634} & 0.600 & \bf0.808 & \bf0.825 & 0.569 & \bf0.766 & \bf0.936 \\
\bf CISR$_{ft}$ & 0.633 & 0.639 & \underline{0.634} & 0.638 & 0.620 & 0.656 & 0.565 & 0.555 & 0.865 \\
\bf LUAR$_{ft}$ & \bf0.764 & \bf0.734 & \bf0.688 & \underline{0.737} & \underline{0.800} & \underline{0.761} & \underline{0.622} & \underline{0.685} & \underline{0.909} \\\bottomrule
\end{tabular}%
}
\caption{\small Bootstrapped test performance (\textbf{AUC}) across all fine-tuned ($_{ft}$) models for the BBN encoding across all distribution combinations. Best performance per combination is bolded and second best, underlined, the differences of which are all statistically significant ($p<0.05$) except for ties. The largest standard error is 0.0004.}
\label{table:ftbbn}
\vspace{-3pt}
\end{table*}

\begin{figure}[!h]\vspace{-9pt}
    \includegraphics[width=.48\textwidth]{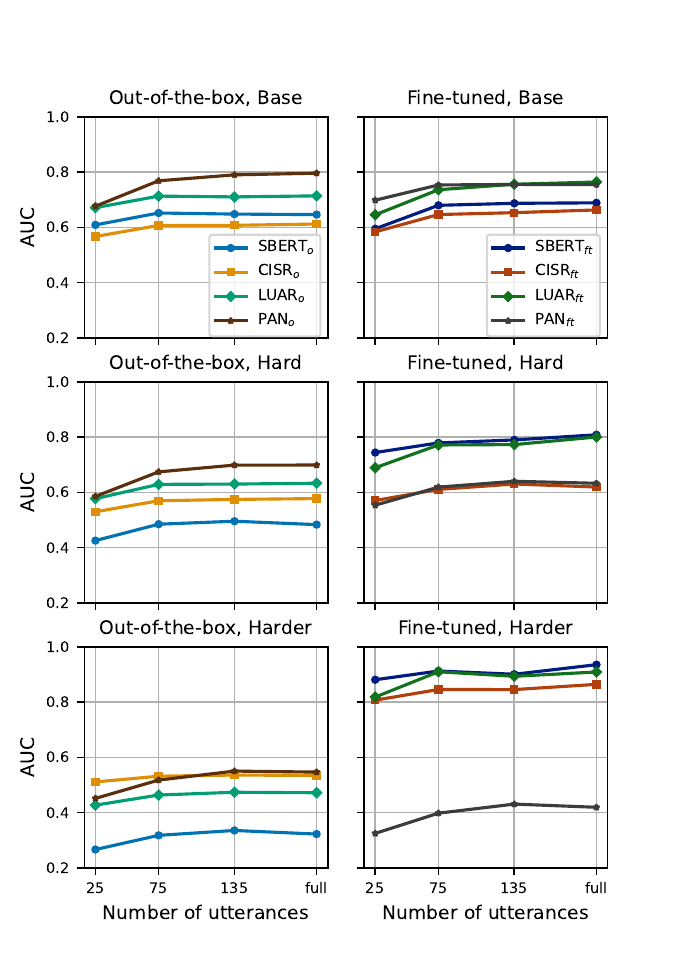}
\caption{\small Bootstrapped AUC test performance (y-axis) across out-of-the-box and fine-tuned models (columns) on the \textbf{BBN} encoding at the 3 levels of difficulty (rows) with the number of utterances per speaker varied (x-axis). Increasing the number of utterances improves performance for all models, with the best generally achieved by 135 utterances.}
\label{fig:uttsbbn}
\vspace{-9pt}
\end{figure}

\begin{table*}[p]
\centering
\resizebox{\linewidth}{!}{%
\small
\begin{tabular}{lccc|ccc|ccc}
\emph{\textbf{EER}} & \multicolumn{3}{c}{\emph{BBN encoding}} & \multicolumn{3}{c}{\emph{LDC encoding}} & \multicolumn{3}{c}{\emph{NormLDC encoding}}\\
\toprule
\bf Model & \bf Base & \bf Hard & \bf Harder & \bf Base & \bf Hard & \bf Harder & \bf Base & \bf Hard & \bf Harder \\ \midrule
\bf TF-IDF$_o$ & \underline{0.307} & \underline{0.377} & 0.502 & 0.305 & 0.372 & 0.495 & 0.286 & \underline{0.349} & 0.479  \\ 
\bf PANgrams$_o$ & \bf0.280 & \bf0.344 & \bf0.475 & \bf0.268 & \bf0.340 & 0.457 & \underline{0.284} & 0.354 & \bf0.466  \\ 
\bf AdHominem$_o$ & 0.459 & 0.466 & 0.512 & 0.428 & 0.444 & 0.471 & 0.449 & 0.442 & 0.484  \\ 
\bf SBERT$_o$ & 0.396 & 0.508 & 0.637 & 0.395 & 0.527 & 0.663 & 0.421 & 0.474 & 0.802  \\ 
\bf CISR$_o$ & 0.422 & 0.448 & \underline{0.481} & 0.374 & 0.380 & \bf0.399 & 0.424 & 0.468 & 0.507  \\ 
\bf LUAR$_o$ & 0.340 & 0.407 & 0.517 & \underline{0.276} & \underline{0.345} & \underline{0.456} & \bf0.240 & \bf0.336 & \underline{0.472}  \\  \midrule
\bf TF-IDF$_{ft}$ & 0.478 & 0.443 & 0.470 & 0.480 & 0.439 & 0.484 & 0.478 & 0.434 & 0.478  \\ 
\bf PANgrams$_{ft}$ & \bf0.300 & 0.405 & 0.554 & \underline{0.308} & 0.406 & 0.568 & \underline{0.285} & \underline{0.416} & 0.569  \\ 
\bf AdHominem$_{ft}$ & 0.454 & 0.458 & 0.438 & 0.434 & 0.448 & 0.438 & 0.423 & 0.450 & 0.473  \\ 
\bf SBERT$_{ft}$ & \underline{0.372} & \bf0.258 & \bf0.142 & 0.358 & \underline{0.256} & \bf0.147 & 0.406 & 0.461 & \bf0.214  \\ 
\bf CISR$_{ft}$ & 0.375 & 0.416 & 0.215 & 0.340 & 0.401 & 0.294 & 0.416 & 0.469 & \underline{0.435}  \\ 
\bf LUAR$_{ft}$ & \bf0.300 & \underline{0.271} & \underline{0.173} & \bf0.230 & \bf0.212 & \underline{0.185} & \bf0.239 & \bf0.215 & \bf0.213  \\ 
\bottomrule
\end{tabular}%
}
\caption{\small Bootstrapped test performance (\textbf{EER}) across all out-of-the-box ($_o$) models and fine-tuned ($_{ft}$) models in the train-test match setting for the BBN, LDC, and NormLDC encodings across all difficulty levels. Best performance for each encoding and difficulty level within $_o$ and $_{ft}$ models (separately) is bolded and second best, underlined, the differences of which are all statistically significant ($p<0.002$) except for ties.}
\label{table:eer}
\vspace{-2.5pt}
\end{table*}

\begin{table*}[t]
\centering
\resizebox{\linewidth}{!}{%
\small
\begin{tabular}{lccc|ccc|ccc}
\emph{\textbf{AUC}} & \multicolumn{3}{c}{\emph{BBN encoding}} & \multicolumn{3}{c}{\emph{LDC encoding}} & \multicolumn{3}{c}{\emph{NormLDC encoding}}\\
\toprule
\bf Model & \bf Base & \bf Hard & \bf Harder & \bf Base & \bf Hard & \bf Harder & \bf Base & \bf Hard & \bf Harder \\ \midrule
\bf LUAR$_o$ & 0.714 & 0.633 & 0.472 & 0.803 & 0.711 & 0.547 & 0.837 & 0.722 & 0.543 \\
\bf LUARnorm$_o$ & 0.717 & 0.614 & 0.439 & 0.794 & 0.682 & 0.490 & 0.831 & 0.704 & 0.524 \\
\bf PreTrain-BBN$_o$ & \underline{0.879} & 0.703 & 0.387 & 0.877 & 0.703 & 0.486 & 0.887 & 0.709 & 0.505 \\ 
\bf PreTrain-LDC$_o$ & 0.821 & 0.580 & 0.342 & \underline{0.905} & 0.707 & 0.406 & \underline{0.906} & 0.699 & 0.418  \\
  \midrule
\bf LUAR$_{ft}$ & 0.764 & 0.801 & 0.909 & 0.844 & 0.872  & 0.894 & 0.844 & 0.869 & 0.876 \\
\bf LUARnorm$_{ft}$ & 0.796 & 0.819 & 0.893 & 0.850 & 0.860 & 0.907 & 0.864 & 0.875 & 0.907 \\ 
\bf PreTrain-BBN$_{ft}$ & \bf0.896 & \bf0.897 & \bf0.932 & 0.899 & \underline{0.902} & \bf0.946 & \underline{0.906} & \underline{0.909} & \bf0.952 \\
\bf PreTrain-LDC$_{ft}$ & 0.877 & \underline{0.894} & \underline{0.930} & \bf0.910 & \bf0.915 & \underline{0.943} & \bf0.909 & \bf0.921 & \underline{0.935} \\
\bottomrule
\end{tabular}%
}
\caption{\small Bootstrapped test performance (\textbf{AUC}) across all LUAR out-of-the-box ($_o$) models and fine-tuned ($_{ft}$) models in the train-test match setting for the BBN, LDC, and NormLDC encodings across all difficulty levels. Best performance for each encoding and difficulty level across $_o$ and $_{ft}$ models (together) is bolded and second best, underlined, the differences of which are all statistically significant ($p<0.001$) except for ties. The largest standard error is 0.0004.}
\label{table:luarall}
\vspace{-2.5pt}
\end{table*}

\begin{table*}[t]
\centering
\resizebox{\linewidth}{!}{%
\small
\begin{tabular}{lccc|ccc|ccc}
\emph{\textbf{EER}} & \multicolumn{3}{c}{\emph{BBN encoding}} & \multicolumn{3}{c}{\emph{LDC encoding}} & \multicolumn{3}{c}{\emph{NormLDC encoding}}\\
\toprule
\bf Model & \bf Base & \bf Hard & \bf Harder & \bf Base & \bf Hard & \bf Harder & \bf Base & \bf Hard & \bf Harder \\ \midrule 
\bf LUAR$_o$ & 0.340 & 0.407 & 0.517 & 0.276 & 0.345 & 0.456 & 0.240 & 0.336 & 0.472  \\ 
\bf LUARnorm$_o$ & 0.344 & 0.420 & 0.547 & 0.276 & 0.364 & 0.503 & 0.252 & 0.356 & 0.485  \\  
\bf PreTrain-BBN$_o$ & 0.200 & 0.363 & 0.580 & 0.201 & 0.359 & 0.500 & 0.196 & 0.346 & 0.489  \\ 
\bf PreTrain-LDC$_o$ & 0.254 & 0.451 & 0.627 & \underline{0.175} & 0.350 & 0.583 & \underline{0.176} & 0.363 & 0.578  \\  \midrule
\bf LUAR$_{ft}$ & 0.300 & 0.271 & 0.173 & 0.230 & 0.212 & 0.185 & 0.239 & 0.215 & 0.213  \\ 
\bf LUARnorm$_{ft}$ & 0.284 & \underline{0.241} & 0.197 & 0.229 & 0.226 & 0.175 & 0.218 & 0.211 & 0.184  \\ 
\bf PreTrain-BBN$_{ft}$ & \bf0.188 & \bf0.181 & \underline{0.152} & 0.181 & \underline{0.175} & \bf0.128 & \underline{0.175} & \underline{0.172} & \bf0.118  \\ 
\bf PreTrain-LDC$_{ft}$ & \underline{0.207} & \bf0.180 & \bf0.150 & \bf0.168 & \bf0.167 & \underline{0.131} & \bf0.174 & \bf0.157 & \underline{0.144}  \\ 
\bottomrule
\end{tabular}%
}
\caption{\small Bootstrapped test performance (\textbf{EER}) across all LUAR out-of-the-box ($_o$) models and fine-tuned ($_{ft}$) models in the train-test match setting for the BBN, LDC, and NormLDC encodings across all difficulty levels. Best performance for each encoding and difficulty level across $_o$ and $_{ft}$ models (together) is bolded and second best, underlined, the differences of which are all statistically significant ($p<0.001$) except for ties.}
\label{table:eerluar}
\vspace{-3pt}
\end{table*}

%% file: main-6291-Aggazzotti.bbl
\begin{thebibliography}{33}
\expandafter\ifx\csname natexlab\endcsname\relax\def\natexlab#1{#1}\fi

\bibitem[{Baayen et~al.(2002)Baayen, Van~Halteren, Neijt, and Tweedie}]{baayen2002}
Harald Baayen, Hans Van~Halteren, Anneke Neijt, and Fiona Tweedie. 2002.
\newblock \href {https://citeseerx.ist.psu.edu/document?repid=rep1&type=pdf&doi=5614de7cd848649bfadaa47adbb951a6529d0915} {An experiment in authorship attribution}.
\newblock In \emph{6$^{es}$ Journ\'ees internationales d'Analyse statistique des Donn\'ees Textuelles (JADT)}, volume~1, pages 69--75. Citeseer.

\bibitem[{Bevendorff et~al.(2020)Bevendorff, Ghanem, Giachanou, Kestemont, Manjavacas, Markov, Mayerl, Potthast, Rangel~Pardo, Rosso, Specht, Stamatatos, Stein, Wiegmann, and Zangerle}]{PAN2020}
Janek Bevendorff, Bilal Ghanem, Anastasia Giachanou, Mike Kestemont, Enrique Manjavacas, Ilia Markov, Maximilian Mayerl, Martin Potthast, Francisco Rangel~Pardo, Paolo Rosso, Guenther Specht, Efstathios Stamatatos, Benno Stein, Matti Wiegmann, and Eva Zangerle. 2020.
\newblock \href {https://doi.org/10.1007/978-3-030-58219-7_25} {Overview of {PAN} 2020: Authorship verification, celebrity profiling, profiling fake news spreaders on {Twitter}, and style change detection}.
\newblock In \emph{Experimental IR Meets Multilinguality, Multimodality, and Interaction}, pages 372--383. Springer International Publishing.

\bibitem[{Boenninghoff et~al.(2019)Boenninghoff, Hessler, Kolossa, and Nickel}]{boenninghoff2019}
Benedikt Boenninghoff, Steffen Hessler, Dorothea Kolossa, and Robert~M. Nickel. 2019.
\newblock \href {https://doi.org/10.1109/BigData47090.2019.9005650} {Explainable authorship verification in social media via attention-based similarity learning}.
\newblock In \emph{IEEE International Conference on Big Data (Big Data)}, pages 36--45.

\bibitem[{Cieri et~al.(2004)Cieri, Graff, Kimball, Miller, and Walker}]{fisher}
Christopher Cieri, David Graff, Owen Kimball, Dave Miller, and Kevin Walker. 2004.
\newblock \href {https://doi.org/https://doi.org/10.35111/w4bk-9b14} {The {Fisher Corpus: A} resource for the next generations of speech-to-text}.

\bibitem[{Danescu-Niculescu-Mizil et~al.(2011)Danescu-Niculescu-Mizil, Gamon, and Dumais}]{danescu}
Cristian Danescu-Niculescu-Mizil, Michael Gamon, and Susan Dumais. 2011.
\newblock \href {https://doi.org/10.1145/1963405.1963509} {Mark my words! {L}inguistic style accommodation in social media}.
\newblock In \emph{Proceedings of the 20th International Conference on World Wide Web}, pages 745--754.

\bibitem[{{Daubert v. Merrell Dow Pharmaceuticals, Inc.}(1993)}]{daubert}
{Daubert v. Merrell Dow Pharmaceuticals, Inc.} 1993.
\newblock 509 U.S. 579.
\newblock \href {https://supreme.justia.com/cases/federal/us/509/579/} {[link]}.

\bibitem[{Ding et~al.(2019)Ding, Fung, Iqbal, and Cheung}]{ding2019}
Steven H.~H. Ding, Benjamin C.~M. Fung, Farkhund Iqbal, and William~K. Cheung. 2019.
\newblock \href {https://doi.org/10.1109/TCYB.2017.2766189} {Learning stylometric representations for authorship analysis}.
\newblock \emph{IEEE Transactions on Cybernetics}, 49(1):107--121.

\bibitem[{Duncan(1974)}]{duncan1974}
Starkey Duncan. 1974.
\newblock \href {https://doi.org/10.1017/S0047404500004322} {On the structure of speaker-auditor interaction during speaking turns}.
\newblock \emph{Language in Society}, 3(2):161--180.

\bibitem[{Fang et~al.(2019)Fang, Wang, Yamagishi, Echizen, Todisco, Evans, and Bonastre}]{fang2019}
Fuming Fang, Xin Wang, Junichi Yamagishi, Isao Echizen, Massimiliano Todisco, Nicholas Evans, and Jean-Fran\c{c}ois Bonastre. 2019.
\newblock \href {https://doi.org/10.48550/arXiv.1905.13561} {Speaker anonymization using x-vector and neural waveform models}.
\newblock eess.AS/1905.13561v1.

\bibitem[{Giles et~al.(2023)Giles, Edwards, and Walther}]{giles2023}
Howard Giles, America~L. Edwards, and Joseph~B. Walther. 2023.
\newblock \href {https://doi.org/10.1016/j.langsci.2023.101571} {Communication accommodation theory: Past accomplishments, current trends, and future prospects}.
\newblock \emph{Language Sciences}, 99.

\bibitem[{Gold and French(2019)}]{gold-french2019}
Erica Gold and John~Peter French. 2019.
\newblock \href {https://eprints.whiterose.ac.uk/150862/} {International practices in forensic speaker comparisons: {S}econd survey}.
\newblock \emph{International Journal of Speech, Language and the Law}, 26:1--20.

\bibitem[{Goldstein-Stewart et~al.(2009)Goldstein-Stewart, Winder, and Sabin}]{goldstein2009}
Jade Goldstein-Stewart, Ransom Winder, and Roberta~Evans Sabin. 2009.
\newblock \href {https://aclanthology.org/E09-1039.pdf} {Person identification from text and speech genre samples}.
\newblock In \emph{Proceedings of the 12th Conference of the European Chapter of the ACL (EACL 2009)}, pages 336--344. Association for Computational Linguistics.

\bibitem[{He et~al.(2013)He, Barbosa, and Kondrak}]{he2023-speakerID-novels}
Hua He, Denilson Barbosa, and Grzegorz Kondrak. 2013.
\newblock \href {https://aclanthology.org/P13-1129} {Identification of speakers in novels}.
\newblock In \emph{Proceedings of the 51st Annual Meeting of the Association for Computational Linguistics (Volume 1: Long Papers)}, pages 1312--1320, Sofia, Bulgaria. Association for Computational Linguistics.

\bibitem[{Kestemont et~al.(2012)Kestemont, Luyckx, Daelemans, and Crombez}]{kestemont2012}
Mike Kestemont, Kim Luyckx, Walter Daelemans, and Thomas Crombez. 2012.
\newblock \href {https://doi.org/https://doi.org/10.1080/0013838X.2012.668793} {Cross-genre authorship verification using unmasking}.
\newblock \emph{English Studies}, 93(3):340--356.

\bibitem[{Khan et~al.(2021)Khan, Fleming, Schofield, Bishop, and Andrews}]{khan2021}
Aleem Khan, Elizabeth Fleming, Noah Schofield, Marcus Bishop, and Nicholas Andrews. 2021.
\newblock \href {https://doi.org/10.18653/v1/2021.naacl-main.415} {A deep metric learning approach to account linking}.
\newblock In \emph{Proceedings of the 2021 Conference of the North American Chapter of the Association for Computational Linguistics: Human Language Technologies}, pages 5275--5287, Online. Association for Computational Linguistics.

\bibitem[{Kimball et~al.(n.d.)Kimball, Iyer, Kao, Colthurst, and Makhoul}]{kimball}
Owen Kimball, Rukmini Iyer, Chia-lin Kao, Thomas Colthurst, and John Makhoul. n.d.
\newblock \href {https://slideplayer.com/slide/5941928/} {Quick transcription of {F}isher data with {WordWave}}.

\bibitem[{Lewis(1997)}]{reuters}
David~D. Lewis. 1997.
\newblock \href {https://kdd.ics.uci.edu/databases/reuters21578/reuters21578.html} {Reuters-21578 text categorization test collection, {D}istribution 1.0}.
\newblock AT\&T Labs-Research.

\bibitem[{Mosteller and Wallace(1964)}]{mostellerwallace}
Frederick Mosteller and David~L. Wallace. 1964.
\newblock \emph{Inference and Disputed Authorship: The Federalist}.
\newblock Addison-Wesley, Reading, MA.

\bibitem[{Najafi and Tavan(2022)}]{najafi-tavan2022}
Maryam Najafi and Ehsan Tavan. 2022.
\newblock \href {https://ceur-ws.org/Vol-3180/paper-215.pdf} {Text-to-text transformer in authorship verification via stylistic and semantical analysis}.
\newblock In \emph{Notebook for PAN at CLEF 2022}, CLEF 2022–Conference and Labs of the Evaluation Forum.

\bibitem[{Pardo et~al.(2022)Pardo, Pellegrino, Dellwo, and M{\"o}bius}]{pardo2022}
Jennifer~S. Pardo, Elisa Pellegrino, Volker Dellwo, and Bernd M{\"o}bius. 2022.
\newblock \href {https://doi.org/10.1016/j.wocn.2022.101196} {Vocal accommodation in speech communication}.
\newblock \emph{Journal of Phonetics}, 95.

\bibitem[{Reimers and Gurevych(2019)}]{sbert}
Nils Reimers and Iryna Gurevych. 2019.
\newblock \href {https://aclanthology.org/D19-1410.pdf} {Sentence-{BERT}: Sentence embeddings using {Siamese BERT-Networks}}.
\newblock pages 3982--3992. Association for Computational Linguistics.

\bibitem[{Rivera-Soto et~al.(2021)Rivera-Soto, Miano, Ordonez, Chen, Khan, Bishop, and Andrews}]{rivera-soto2021}
Rafael~A. Rivera-Soto, Olivia~Elizabeth Miano, Juanita Ordonez, Barry~Y. Chen, Aleem Khan, Marcus Bishop, and Nicholas Andrews. 2021.
\newblock \href {https://aclanthology.org/2021.emnlp-main.70} {Learning universal authorship representations}.
\newblock In \emph{Proceedings of the 2021 Conference on Empirical Methods in Natural Language Processing}, pages 913--919. Association for Computational Linguistics.

\bibitem[{Sacks(1992)}]{sacks1992}
Harvey Sacks. 1992.
\newblock \emph{Lectures on Conversation}, volume~1.
\newblock Blackwell.

\bibitem[{Scheijen(2020)}]{scheijen2020}
Nelleke Scheijen. 2020.
\newblock \href {http://resolver.tudelft.nl/uuid:100073ef-bb5b-42a0-a957-70d2e7916178} {Forensic speaker recognition: {B}ased on text analysis of transcribed speech fragments}.
\newblock Master's thesis, Delft University of Technology.

\bibitem[{Sisman et~al.(2021)Sisman, Yamagishi, King, and Li}]{sisman2021}
Berrak Sisman, Junichi Yamagishi, Simon King, and Haizhou Li. 2021.
\newblock \href {https://doi.org/10.1109/TASLP.2020.3038524} {An overview of voice conversion and its challenges: {F}rom statistical modeling to deep learning}.
\newblock \emph{IEEE/ACM Transactions on Audio, Speech, and Language Processing}, 29:132--157.

\bibitem[{Snyder et~al.(2018)Snyder, Garcia-Romero, Sell, Povey, and Khudanpur}]{xvectors2018}
David Snyder, Daniel Garcia-Romero, Gregory Sell, Daniel Povey, and Sanjeev Khudanpur. 2018.
\newblock \href {https://doi.org/10.1109/ICASSP.2018.8461375} {X-vectors: Robust {DNN} embeddings for speaker recognition}.
\newblock In \emph{2018 IEEE International Conference on Acoustics, Speech and Signal Processing (ICASSP)}, pages 5329--5333.

\bibitem[{Stamatatos(2018)}]{stamatatos2018}
Efstathios Stamatatos. 2018.
\newblock \href {https://doi.org/https://doi.org/10.1002/asi.23968} {Masking topic-related information to enhance authorship attribution}.
\newblock \emph{Journal of the Association for Information Science and Technology}, 69(3):461--473.

\bibitem[{Stamatatos et~al.(2023)Stamatatos, Kredens, Pezik, Heini, Bevendorff, Stein, and Potthast}]{pan2023}
Efstathios Stamatatos, Krzysztof Kredens, Piotr Pezik, Annina Heini, Janek Bevendorff, Benno Stein, and Martin Potthast. 2023.
\newblock \href {https://ceur-ws.org/Vol-3497/paper-199.pdf} {Overview of the authorship verification task at {PAN} 2023}.
\newblock In \emph{CLEF 2023: Conference and Labs of the Evaluation Forum}.

\bibitem[{Tripto et~al.(2023)Tripto, Uchendu, Le, Setzu, Giannotti, and Lee}]{hansen2023}
Nafis~Irtiza Tripto, Adaku Uchendu, Thai Le, Mattia Setzu, Fosca Giannotti, and Dongwon Lee. 2023.
\newblock \href {http://arxiv.org/abs/2310.16746} {{HANSEN: H}uman and {AI} spoken text benchmark for authorship analysis}.
\newblock cs.CL/2310.16746v1.

\bibitem[{Wang et~al.(2023)Wang, Aggazzotti, Kotula, Soto, Bishop, and Andrews}]{wang2023}
Andrew Wang, Cristina Aggazzotti, Rebecca Kotula, Rafael~Rivera Soto, Marcus Bishop, and Nicholas Andrews. 2023.
\newblock \href {https://doi.org/10.1162/tacl_a_00610} {{Can Authorship Representation Learning Capture Stylistic Features?}}
\newblock \emph{Transactions of the Association for Computational Linguistics}, 11:1416--1431.

\bibitem[{Watt and Brown(2020)}]{watt-brown2020}
Dominic Watt and Georgina Brown. 2020.
\newblock \href {https://doi.org/https://doi.org/10.4324/9780429030581} {\emph{Forensic phonetics and automatic speaker recognition: {The} complementarity of human- and machine-based forensic speaker comparison}}, chapter~25. Routledge.

\bibitem[{Wegmann et~al.(2022)Wegmann, Schraagen, and Nguyen}]{wegmann2022}
Anna Wegmann, Marijn Schraagen, and Dong Nguyen. 2022.
\newblock \href {https://aclanthology.org/2022.repl4nlp-1.26} {Same author or just same topic? {T}owards content-independent style representations}.
\newblock In \emph{Proceedings of the 7th Workshop on Representation Learning for NLP}, pages 249--268. Association for Computational Linguistics.

\bibitem[{Zhu and Jurgens(2021)}]{zhu2021}
Jian Zhu and David Jurgens. 2021.
\newblock \href {https://doi.org/10.18653/v1/2021.emnlp-main.25} {Idiosyncratic but not arbitrary: Learning idiolects in online registers reveals distinctive yet consistent individual styles}.
\newblock In \emph{Proceedings of the 2021 Conference on Empirical Methods in Natural Language Processing}, pages 279--297. Association for Computational Linguistics.

\end{thebibliography}
